\documentclass[letterpaper, 10 pt, conference]{ieeeconf}  

\usepackage{tabularx}
\usepackage{hyperref}

\IEEEoverridecommandlockouts                              

\usepackage{graphicx} 
\usepackage{amsmath} 
\usepackage{amssymb}  

\title{\LARGE \bf
A Production Scheduling Framework for Reinforcement Learning Under Real-World Constraints
}
\author{Jonathan Hoss$^{1}$, Felix Schelling$^{1}$ and Noah Klarmann$^{1}$%
\thanks{*The work has been performed in the Cynergy4MIE project (GA. 101140226)}%
\thanks{$^{1}$ Faculty of Management and Engineering, Rosenheim Technical University of Applied Sciences, Germany, Correspondence: Jonathan Hoss, {\tt\small jonathan.hoss@th-rosenheim.de}}%
}  

\begin{document}





\maketitle
\thispagestyle{empty}
\pagestyle{empty}

\begin{abstract}
The classical Job Shop Scheduling Problem (JSSP) focuses on optimizing makespan under deterministic constraints. Real-world production environments introduce additional complexities that cause traditional scheduling approaches to be less effective. Reinforcement learning (RL) holds potential in addressing these challenges, as it allows agents to learn adaptive scheduling strategies. However, there is a lack of a comprehensive, general-purpose frameworks for effectively training and evaluating RL agents under real-world constraints. To address this gap, we propose a modular framework that extends classical JSSP formulations by incorporating key \mbox{real-world} constraints inherent to the shopfloor, including transport logistics, buffer management, machine breakdowns, setup times, and stochastic processing conditions, while also supporting multi-objective optimization. The framework is a customizable solution that offers flexibility in defining problem instances and configuring simulation parameters, enabling adaptation to diverse production scenarios. A standardized interface ensures compatibility with various RL approaches, providing a robust environment for training RL agents and facilitating the standardized comparison of different scheduling methods under dynamic and uncertain conditions. We release JobShopLab as an open-source tool for both research and industrial applications, accessible at: \texttt{https://github.com/proto-lab-ro/jobshoplab}
\end{abstract}

\section{INTRODUCTION}


The Job Shop Scheduling Problem (JSSP) is an optimization problem in production planning and control that involves scheduling a set of jobs on a limited number of machines. Each job comprises a predetermined sequence of operations, with each operation requiring specific machinery and processing times. The objective of classical JSSP is typically to minimize the makespan, the total time required to complete all jobs. \cite{pinedoSchedulingTheoryAlgorithms2022} The complexity of JSSP stems from the constraints that machines can process only one operation at a time and that the sequence of operations must be strictly adhered to. Its nature is characterized by its NP-hard classification, implying that the computational effort required to solve them increases exponentially with the size of the problem. This makes them particularly challenging to address optimally for large instances.

The demands for approaches to JSSP arise from its essential role in short-term planning within the production planning and control process, as classified by Jodlbauer et al. \cite{jodlbauerProduktionsoptimierungWertschaffendeSowie2008}. It is applied to real-time execution, resource allocation, and the sequencing of operations while managing dynamic disruptions. These disruptions introduce general requirements for flexibility to accommodate unforeseen events, robustness to ensure operational stability, dynamism for continuous adaptation to changing conditions, and the balancing of multiple optimization objectives. While various approaches for solving JSSP have proven effective in structured and predictable environments, challenges emerge when applied to real-world scenarios with uncertainties \cite{pinedoSchedulingTheoryAlgorithms2022}.

The approaches for optimizing the JSSP are generally classified into two categories: exact methods, which aim to find the global optimum, and approximate methods, which provide solutions that are near-optimal. While exact methods strive for an optimal solution, they are computationally intensive and may not be feasible for large-scale or complex problems when seeking a solution within a reasonable timeframe.  \cite{zhangReviewJobShop2019} Approximate methods, on the other hand, aim to deliver solutions that are both computationally efficient and of acceptable quality within a reasonable timeframe. Approximate methods include search techniques or heuristic approaches, such as priority dispatching rules (PDR). 

Reinforcement learning (RL) is considered an approximate method and shows significant potential, particularly for complex scenarios. Although RL training can be resource-intensive, once trained, RL agents can be deployed efficiently, enabling real-time applications with minimal computational overhead. RL offers a distinct advantage over heuristic techniques by allowing agents to learn adaptive scheduling policies through interactions with dynamic environments. As Pinedo \cite{pinedoSchedulingTheoryAlgorithms2022} notes regarding traditional JSSP methods, "the more randomness the system is subject to, the simpler the scheduling rules ought to be." This observation underscores the limitations of deterministic models and highlights the potential of \mbox{RL-based} approaches. However, an agent's ability to perform effectively in real-world scenarios depends on the accuracy of the environment in simulating the complexities of such conditions. Many RL-based approaches abstract real-world challenges to maintain comparability with traditional methods, which hinders their practical application in industrial contexts. Existing frameworks vary in complexity and often focus on simplified objectives rather than adopting a multi-objective perspective. A broader approach is essential in modern manufacturing, where adaptability, efficiency, and sustainability are paramount.

This creates a need for a framework that accurately simulates real-world scenarios, accounting for diverse constraints and uncertainties in real-world production environments. This paper addresses this gap by proposing a framework that provides a solid foundation for training and evaluating RL agents, enabling consistent benchmarking of RL-based scheduling solutions.

The main contributions of this paper are as follows:
\begin{itemize}
    \item We introduce key real-world extensions to the classical JSSP and demonstrate how the real-world job shop environment can be modeled.
    \item We present a comprehensive framework with a modular architecture that enables the representation of various production environments, featuring a standardized interface for RL agents and offering flexible configuration and specification options, while also providing a solution that is extensible to custom problems and features.
    \item We demonstrate the functionality of the proposed framework for both training and benchmarking RL agents, providing a proof of concept that supports its practical application in both research and industry settings.
\end{itemize}

\section{Literature Review}
RL is a machine learning approach where an agent learns optimal actions through trial and error by interacting with an environment and receiving feedback in the form of reward \cite{suttonReinforcementLearningIntroduction2018}.


The application of RL to the JSSP has seen significant growth in recent years. However, despite promising advances, existing approaches frequently fail to capture the complexities of real-world production environments. In the literature on RL applications for JSSP, it is apparent that the field can largely be divided into two areas: solutions focused on the classical JSSP, aiming to determine an optimal methodology for problem-solving, and solutions attempting to extend the classical JSSP to specific applications in order to get better results when transferring the agent to a real-world scenario. This section provides a critical overview of the literature, categorizing contributions into classical JSSP solutions, extended JSSPs, and an evaluation on available frameworks in this area. 

\subsection{Classical RL Solutions for JSSP}
The classical JSSP, as referred to in this paper, describes a JSSP with the sole objective of minimizing makespan for a given number of jobs to machines without additional complexity.
Various studies have applied RL to address the classical JSSP. Tassel et al. \cite{tasselReinforcementLearningEnvironment2021} contributed to the field by formulating the JSSP as a single-agent problem where the agent's action consists of selecting the next job for processing. Rather than defining the minimal makespan as a sparse reward, they introduced a dense reward that accounts for machine idle times, defined as the times when machines are not working on any jobs. This approach, when combined with a state representation optimized via Proximal Policy Optimization (PPO), outperforms traditional job shop scheduling methods.
Zhao and Zhang \cite{zhaoEndtoEndDeepReinforcement2022} and Tassel et al. \cite{tasselReinforcementLearningDispatching2022} evaluate a different approach by defining PDRs, rather than job selection, as the action space. This modification simplifies decision-making while yielding high returns and improving generalization across different scheduling scenarios. 
Qiao et al. \cite{qiaoOptimizationJobShop2024} employ a deep RL algorithm with a disjunctive graph state model and convolutional networks for action selection, further enhancing adaptability to varied problem instances.
Altenmüller et al. \cite{altenmullerReinforcementLearningIntelligent2020} apply Q-learning combined with discrete-event simulation to improve order dispatching in job shops with strict time constraints. Their approach rewards urgency-based scheduling decisions and demonstrated superior performance in minimizing time constraint violations compared to traditional heuristics. However, the work focuses on single-objective optimization without considering broader production factors.

The papers mentioned present solutions demonstrating that RL can outperform traditional mathematical approaches and heuristics. Yet, Schneevogt et al. \cite{schneevogtOptimizingJobShop2024} highlighted the limitations of such theoretical solutions, which do not adequately account for real-world complexities and therefore limit their applicability in dynamic production settings. They stress the significance of factors such as machine setup times, batch sizes, and intralogistics for real-world applicability. 
Consequently, research has increasingly focused on extending the classical JSSP to incorporate real-world constraints, as discussed in the following section.

\subsection{Extensions for Real-World Constraints}
Extensions of the classical JSSP start with targeted modifications that address specific real-world constraints. For instance, Chang et al. \cite{changDeepReinforcementLearning2022} introduced setup times for machine operations, significantly improving schedule feasibility. This demonstrates how increasing problem complexity can enhance practical applicability by producing schedules that are more adaptive and realistic. Beyond such foundational modifications, industry- or use case-specific adaptations are also introduced to address particular requirements. In principle, these extensions can be categorized into three key areas: stochasticity, transportation, and multi-objective optimization.

Stochasticity refers to the incorporation of uncertainty into the problem, where certain elements are modeled probabilistically. Liu et al. \cite{liuDeepReinforcementLearning2022} extended the classical JSSP by introducing random job arrivals, where new jobs appear dynamically, along with machine breakdowns and processing time variability. Their approach utilized a distributed and hierarchical architecture with a Double Deep Q-Network for dynamic flexible job shop scheduling, aiming to minimize the makespan while adapting to disruptions. Through reward shaping and specialized state-action representations, their system demonstrated enhanced adaptability and efficiency in numerical simulations, outperforming traditional methods in dynamic environments. Similarly, Wu et al. \cite{wuDeepReinforcementLearning2024} developed a deep RL model to manage job scheduling under uncertain processing times. By incorporating a novel state representation and a reward function based on machine idle time, their approach further improved efficiency and adaptability in dynamic scheduling scenarios.

Another extension of the classical JSSP involves the integration of transport management, which addresses the coordination between job scheduling and the production logistics. Li et al. \cite{liRealtimeDatadrivenDynamic2022} incorporated Automated Guided Vehicles (AGVs) alongside PDRs, proving the importance of addressing transport scheduling to enhance system efficiency. Furthermore, Hu et al. \cite{huDeepReinforcementLearning2020} proposed a Deep RL-based framework for real-time scheduling of AGVs to optimize transport logistics and reduce makespan. 

The third area of extension focuses on leveraging the potential of RL for multi-objective optimization in job shop scheduling. For example, Zhao and Zhang \cite{zhaoEndtoEndDeepReinforcement2022} proposed a multi-objective optimization model that considers makespan alongside minimal total delay to a due time, reflecting the practical requirements of real-world production environments. Loffredo et al. \cite{loffredoReinforcementLearningSustainability2024} developed an RL-based energy-efficient control framework that balanced production throughput and energy consumption in multi-stage production lines, achieving significant energy savings. Wu et al. \cite{wuEfficientMultiObjectiveOptimization2023} proposed a dual-layer deep Q-network framework for dynamic flexible job shop scheduling, optimizing both makespan and job delay time through bi-layer agents tasked with selecting optimization goals and PDRs. 
Some papers combine different extensions to address complex challenges. For example, Yu et al. \cite{yuRobustOptimalScheduling2023} included multi-objective optimization and stochasticity. Their approach, driven by multiple performance objectives like job cycle time, on-time delivery, equipment availability and scheduling robustness, integrated job dispatching and equipment maintenance in a closed-loop adaptive optimization framework. This method showed improved efficiency and resilience to scheduling uncertainties, demonstrating the value of combining multiple objectives in scheduling.

The mentioned approaches highlight RL's potential for addressing dynamic production complexities, including logistics and operational uncertainties, as well as managing multiple objectives. So far, research focuses on narrowly defined JSSP extensions, without addressing the comprehensive range of real-world complexities. To enable research in a holistic approach, a framework capable of representing various interconnected factors and dynamics in production systems is needed. Therefore, the following section focuses on evaluating the frameworks that are currently available.

\subsection{Frameworks for Real-World JSSP}


To identify existing solutions that support RL-based scheduling in dynamic production settings, we analyze frameworks from the literature based on key functional requirements, including modularity, support for real-world constraints, extensibility, multi-objective optimization, stochasticity, flexible action and observation spaces, visualization and analysis tools, as well as reusability and accessibility. We summarize this comparison in Table \ref{tab:framework_comparison} to position our own contribution.



One examined framework is that of Serrano-Ruiz et al. \cite{serrano-ruizJobShopSmart2024}, which provides a tailored environment for evaluating a single RL agent in a JSSP setting. The job shop is modeled with fixed observation and action spaces, buffers, and deterministic multi-machine routing, incorporating partial modularity, stochastic elements, and basic real-world constraints. However, the framework lacks extensibility and flexibility in environment configuration. Its single-agent and single-objective design restrict its adaptability to dynamic production settings and limit its reusability for broader \mbox{RL-based} JSSP research.

Liao et al. \cite{liaoLearningScheduleJobShop2022} propose a hierarchical RL approach using graph neural networks. Their method targets classic JSSP instances and addresses scalability through multiple sub-policies for different scheduling phases. However, the framework does not emphasize modularity, extensibility, multi-objective optimization, or the incorporation of real-world constraints, limiting its applicability to industrial environments.

A framework developed by Kuhnle et al. \cite{kuhnleDesigningAdaptiveProduction2021} implements an RL-based adaptive control system for a real-world job shop dispatching scenario. It incorporates stochasticity and partial modularity, addressing some real-world constraints, but lacks flexibility in action and observation spaces, and visualization tools.

Reijnen et al. \cite{reijnenJobShopScheduling2023} present an open-source benchmark suite covering various scheduling problem variants. While their focus lies on standardized benchmarking across problem types, the framework fails to support multi-objective optimization and offers only limited support for real-world constraints, modularity, extensibility, and visualization.

The reviewed frameworks pursue distinct objectives, such as controlled RL evaluation, domain-specific control design, or the development of benchmark collections. Consequently, they are not intended to serve as general-purpose environments for RL in real-world JSSP. As outlined above, they address selected aspects of the problem but fall short in meeting the broader requirements for modularity, extensibility, or comprehensive real-world constraint modeling. In contrast, our framework is designed for accessibility and reuse. It provides a modular, extensible architecture that captures real-world conditions such as transport delays, machine breakdowns, and stochastic processing times and supports multi-objective optimization as well as flexible action and observation spaces, enabling various RL applications in dynamic, industrial JSSP environments.

\begin{table*}[ht]
    \vspace{10pt}  

    \caption{Comparison of simulation frameworks for RL-based JSSP under real-world constraints}
    \centering
    \begin{tabularx}{\textwidth}{lccccc}
        \hline
        \textbf{Criterion} & \textbf{Serrano-Ruiz et al. \cite{serrano-ruizJobShopSmart2024}}  & \textbf{Liao et al.  \cite{liaoLearningScheduleJobShop2022}} & \textbf{Kuhnle et al. \cite{kuhnleDesigningAdaptiveProduction2021}} & \textbf{Reijnen et al. \cite{reijnenJobShopScheduling2023}} & \textbf{Ours} \\
        \hline
        Modularity & Partial & No & Partial & Partial & Yes \\
        Real-world constraints & Partial & No & Partial & Partial & Comprehensive \\
        Extensibility & No & No & Partial & Partial & Yes \\
        Stochasticity & Partial & Partial & Yes & Partial & Yes \\
        Multi-objective optimization & No & No & Partial & No & Yes \\
        Action space flexibility & No & Partial & Partial & Partial & Yes \\
        Observation space flexibility & No & Partial & Partial & Partial & Yes \\
        Visualization and analysis & No & No & No & Partial & Yes \\
        Reusability and accessibility & Partial & No & No & Yes & Yes \\
        \hline
    \end{tabularx}
    \label{tab:framework_comparison}
\end{table*}

\section{Methods}
In the following section, an overview of the key design considerations for developing the framework is provided. Specifically, we identify and formalize the real-world constraints to extend the classical JSSP and outline the core concept of modeling a realistic job shop environment.

\subsection{Formalizing Real-World Constraints}
The classical JSSP considers a production environment with a finite set of Jobs $\mathcal{J}$. Each job $J_i \in \mathcal{J}$ consists of an ordered sequence of $m$ operations $O_{i,1}, O_{i,2}, \dots, O_{i,m}$, which must be executed sequentially, i.e.
$$
O_{i,1} \prec O_{i,2} \prec \dots \prec O_{i,m} \quad \forall J_i \in \mathcal{J}  \eqno{(1)}
$$
Each operation $O_{i,j}$ is assigned to a machine $M_k \in \mathcal{M}$. It requires an uninterrupted processing time $p_{i,j} > 0$. Machines can process only one operation at a time, so assigned operations must not overlap. The classical optimization objective is to find a schedule that minimizes the makespan, where $Z_{i,j}$ is the starting time of operation $O_{i,j}$.
$$
C_{max} = max_{i,j}(C_{i,j}=Z_{i,j} + p_{i,j})  \eqno{(2)}
$$


This basic formulation addresses fundamental scheduling problems in an idealized job shop, but lacks real-world realism. It assumes seamless job transfers between machines, ignoring transport delays or disruptions, and does not account for events like machine breakdowns.

To better reflect the unpredictable and dynamic nature of production environments, we propose a framework that incorporates extensions designed to model real-world constraints. The following section classifies these extensions using the standard three-field notation $\alpha \mid \beta \mid \gamma$ \cite{pinedoSchedulingTheoryAlgorithms2022}, formalizing the scope and the types of JSSP that can be addressed by the proposed framework.

\begin{itemize}
    \item $\alpha$ defines the machine environment,
    \item $\beta$ specifies problem constraints, and
    \item $\gamma$ denotes optimization objectives.
\end{itemize}

The classical JSSP can be formulated as $J\vert \vert C_{\max}$ without any problem constraints $\beta$.
Based on the requirements of real-world production processes, the proposed framework extends the classical JSSP to include the following problem constraints: transport management, buffer management, machine setup and breakdown modeling, and stochastic processing times.

\textbf{Transport management} is a critical factor in job shop operations, particularly when transport resources are limited or travel distances vary significantly. Incorporating transport management into the scheduling framework supports adaptive strategies for optimizing resource use and reducing delays. It is modeled as follows:

\begin{itemize}
    \item Transport times represent delays when relocating jobs between machines.
    \item Transport capacity reflects the number of units and their ability to carry multiple jobs.
    \item Transport breakdowns account for downtime due to malfunctions or maintenance.
    \item Occupation times during loading, travel, and unloading represent periods when units are unavailable.
\end{itemize}

\textbf{Buffer management} introduces realistic constraints through limited buffer capacity and configurable buffer behavior. Unlike classical JSSP models that assume infinite buffer space, we incorporate the following aspects:

\begin{itemize}
    \item Buffer capacity limits storage space; insufficient capacity can cause congestion.
    \item Buffer pick sequence constraints model whether buffers enforce fixed or arbitrary retrieval orders.
\end{itemize}

\textbf{Machine setup and breakdown modeling} captures setup times and stochastic disruptions that affect machine availability and scheduling efficiency, as follows:

\begin{itemize}
    \item Machine setup times depend on the sequence of job types at a machine.
    \item Machine breakdowns are modeled probabilistically to reflect unplanned downtime.
\end{itemize}

Another factor incorporated into the framework is stochasticity, which is represented by probabilistic modeling of transport times and machine breakdowns. Additionally, processing times at machines are modeled as stochastic variables. In real-world production, processing times vary due to machine wear, operator efficiency, and material inconsistencies. Including stochastic processing times enhances simulation realism and supports the development of more robust scheduling strategies.

Building upon the framework extensions that address problem constraints $\beta$, the framework also incorporates diverse optimization objectives $\gamma$. In real-world JSSP applications, these objectives are inherently multidimensional and extend beyond simple makespan minimization. The proposed framework supports simultaneous optimization of multiple objectives, covering both primary and secondary goals. Primary objectives include minimizing makespan, maximum lateness, total weighted completion time, total weighted tardiness, and the weighted number of tardy jobs, all aimed at ensuring efficient job scheduling and adherence to due dates. Beyond these, secondary objectives address critical production factors such as reducing energy consumption during machine operations, minimizing buffer usage to lower capital commitment, decreasing lead time to improve responsiveness, enhancing flexibility to adapt to disruptions and varying order mixes, and optimizing machine utilization by balancing workloads based on machine importance or cost.

By integrating these constraints and objectives, the proposed framework bridges the gap between the classical JSSP and the complexities of real-world production environments. This enables the development and evaluation of RL agents capable of making robust scheduling decisions under uncertain and dynamic conditions.

\subsection{Modeling the Job Shop Environment}
To represent real-world job shop dynamics, we propose a modular, stateless simulation framework built around interacting sub-state machines. The stateless design ensures flexibility, reproducibility, and modularity: every state fully encodes the production system at a specific simulation time, independent of history, forming a Markov Decision Process where transitions depend solely on the current state and actions.


In this framework, an RL agent functions as a dispatching controller. It receives an observation of the current state, selects actions from a predefined action space, and receives a reward based on the resulting transitions. The agent's actions include scheduling jobs on machines and dispatching transport units. Although individual agent actions may be valid, their combined effects can violate system constraints. Invalid transitions are rejected with an error response, guiding the agent toward feasible decisions. These actions do not directly modify the environment; instead, each action $ a \in \mathcal{A} $ triggers a set of events $ \mathcal{E}_{agent}$, which are handled by the state machine. In addition to agent-triggered events, autonomous events $\mathcal{E}_{\text{auto}}$ occur based on system dynamics (e.g., machine breakdowns).

Each simulation time $ t \geq  0 $ is associated with a set of scheduled events $ \mathcal{E}(t) $, comprising both agent-triggered and autonomous events:

$$
\mathcal{E}(t) = \mathcal{E}_{agent}(t) \cup \mathcal{E}_{\text{auto}}(t) \eqno{(3)}
$$

\begin{figure}
    \includegraphics[width=\linewidth]{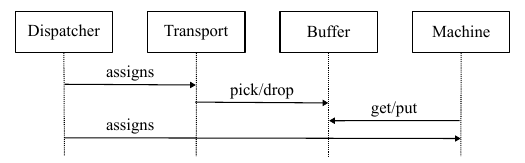}
    \caption{Sequence diagram illustrating interactions between components}
    \vspace{-5mm}  
    \label{Fig:Sequence Diagram}
\end{figure}

The environment uses a modular architecture composed of sub-state machines for transport units, buffers, and machines. Figure~\ref{Fig:Sequence Diagram} illustrates their coordinated interaction, where assignments issued by the dispatcher and physical operations (e.g., pick/drop, get/put) are synchronized to ensure consistent state transitions.

Each internal event $ e \in \mathcal{E}$ corresponds to an atomic transition. The atomic transition function is defined as:
$$
T: \mathcal{S} \times \mathcal{E} \rightarrow \mathcal{S} \eqno{(4)}
$$
Here, $ \mathcal{S} $ denotes the set of valid system states and $ \mathcal{E} $ the set of atomic internal events.

To handle concurrent or dependent events, a priority function $ \rho $ is introduced:
$$
\rho: \mathcal{E} \rightarrow \mathbb{R} \eqno{(5)}
$$

At each simulation step $t$, events are processed in descending order of priority to ensure casual consistency. 
$$
(e_1, e_2, \dots , e_n) = sortDescending(\mathcal{E}(t), \rho) \eqno{(6)}
$$
For instance, transport deliveries precede machine starts, and machine completions precede transport pickups when occurring simultaneously.

Furthermore, breakdown events $ B \subset \mathcal{E}(t) $ always take precedence:
$$
\forall b \in B,\quad \forall e \in \mathcal{E} \setminus B,\quad \rho(b) > \rho(e) \eqno{(7)}
$$

The state machine applies the corresponding state transitions sequentially for all events scheduled at time $t$:
$$
s_{t,0} = s_t,\quad s_{t,i} = T(s_{t,i-1}, e_i),\quad i = 1,\dots,n \eqno{(8)}
$$
After processing all events, the updated system state is defined as:
$$
s_{t+1} = s_{t,n} \eqno{(9)}
$$
For notational convenience, we define a composite transition function:
$$
\hat{T}(s_t, \mathcal{E}(t)) := T(T(\cdots T(T(s_t, e_1), e_2), \cdots), e_n) \eqno{(10)}
$$
$$
s_{t+1} = \hat{T}(s_t, \mathcal{E}(t)) \eqno{(11)}
$$

\begin{figure}
    \includegraphics[width=\linewidth]{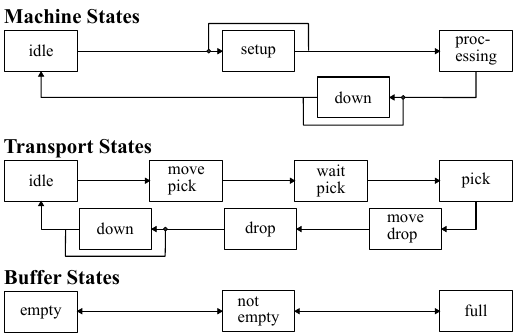}
    \caption{State transition diagram of machine, transport, and buffer components}
        \vspace{-5mm}  
    \label{Fig:Component States}
\end{figure}

The typical state transitions of the core components machines, transport units, and buffers are illustrated in Figure~\ref{Fig:Component States}. These components undergo state changes that reflect their operational behavior during the simulation.

This event-driven, modular approach allows for scalable simulation, accurate modeling of asynchronous operations, and integration with RL-based scheduling policies.

\section{Implementation, Validation, and Verification of the JSSP Framework}
\subsection{Framework Architecture}

\begin{figure}
    \includegraphics[width=\linewidth]{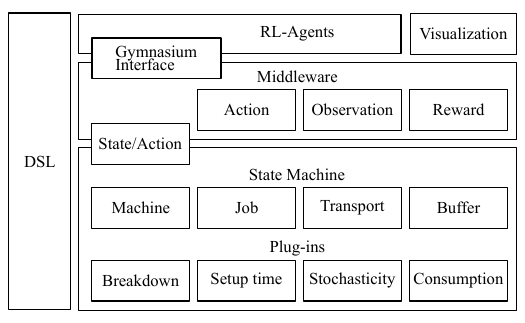}
    \caption{Framework architecture}
    \vspace{-5mm}  
    \label{Fig:Architecture}
\end{figure}

The goal of the proposed framework is to offer a modular, scalable, and adaptable environment for solving real-world JSSP. To achieve this, the architecture is divided into distinct components, as shown in Figure~\ref{Fig:Architecture}, enabling a customizable platform for production process simulation and RL agent training.

The framework architecture is structured as follows:

\begin{itemize}
    \item The state machine serves as the central discrete simulation component, managing the states of the production environment.

    \item The middleware enables communication between the state machine and external components, such as the RL agent, managing tasks like data preprocessing and event triggering. It includes the observation factory, action factory, and reward factory, which define the RL agent’s available observation space, action space, and reward structure for controlling the production process.

    \item The Domain-Specific Language (DSL) provides a structured approach for adapting the framework to a specific JSSP. It consists of a DSL for Problem Instances, which defines key elements to capture the complete dynamic system state, and a DSL for Framework Configuration, which allows customization of observation spaces, reward structures, and action spaces.
\end{itemize}

The architecture emphasizes customizability and modularity, which is why the state machine is designed with a plug-in architecture, allowing for the seamless addition of custom features. A set of plug-ins $\mathcal{P} = \{ P_1, P_2, \dots, P_m \}$ is defined, where each plug-in $P_i$ is modeled as a transformation function: 
$$
P_i : \mathcal{S} \times \mathcal{E} \times \Theta_i \to S \times \Xi_i \eqno{(12)}
$$
where $\Theta_i$ and $\Xi_i$ denote plug-in-specific parameters and auxiliary outputs, respectively.

The following plug-ins are initially implemented:

\begin{itemize}
    \item The breakdown plug-in dynamically simulates machine failures based on predefined probabilities.
    \item The setup time plug-in models setup durations for job transitions on machines. 
    \item The stochasticity plug-in incorporates randomness to reflect real-world variability.
    \item The consumption plug-in tracks resource consumption, including energy usage.
\end{itemize}


To make the framework accessible to a broader audience, the middleware provides a standardized interface for communication with the RL agent. It utilizes the Observation, Reward, and Action implementations from the Gymnasium Framework by Towers et al.~\cite{towersGymnasiumStandardInterface2024}, providing factories that define these components. The framework offers a template for factory definition, allowing users to create custom factories to tailor observation, action, and reward spaces to their research needs.

This modular design inherently supports multi-objective optimization by enabling multiple, configurable reward signals within the framework.

Formally, this mapping can be described as:
$$
(\mathcal{S}, \mathcal{A}) \xrightarrow{\text{Factories}} \{\text{Observation}, \text{Action}, \text{Reward}\} \eqno{(13)}
$$

The initial version of the framework provides the following factory types:

\begin{itemize}
    \item A simple JSSP observation factory generates a basic observation space that captures the status of jobs and machines, providing essential information for RL decision-making.
    \item A binary action factory defines a binary action space where the agent can decide whether to perform an action or not, offering a simplified decision-making process for certain tasks.
    \item A multidiscrete action factory provides a multidimensional action space with discrete options, allowing the agent to select from a range of possible actions based on the system state.
    \item A minimize makespan reward factory supports a reward mechanism aimed at minimizing the total production time, driving the agent to optimize the overall production process.
\end{itemize}

The framework introduces a DSL to provide a general-purpose way of describing complex production scenarios. The DSL enables users to define problem instances and framework configuration in a structured and reusable manner. It is divided into two primary sections: the DSL for Problem Instances and the DSL for Framework Configuration.

The DSL for Problem Instances specifies the production environment’s key characteristics, providing the foundational data for simulation and control. It defines the environment where production occurs and ensures the system’s dynamic state is fully captured. This includes:

\begin{itemize}
    \item Jobs: Detailed descriptions of production orders, including parameters such as processing times, job types, and priorities.
    \item Machines: Characteristics of available machines, including specifications like processing speeds, capacity, and maintenance states.
    \item Transport Systems: Definitions of conveyors, AGVs, and other transport systems, along with their operational states.
    \item States: The current status of jobs, machines, and transport systems. From this information, the complete system state is generated and continuously updated throughout the simulation.
\end{itemize}

The DSL for Framework Configuration provides a structured way to define how RL agents perceive, act, and learn within the simulated production environment. It allows users to specify the observation space, action space, and reward function in a declarative manner, enabling flexible configuration of Observation, Action, and Reward Factories.


Moreover, the framework incorporates a 3D visualization tool constructed with three.js, enabling monitoring and replayable, scene-by-scene animations of the scheduling process for improved analysis and evaluation.

\subsection{Proof of Concept} 
To validate and verify the proposed framework, we trained an RL agent within our simulation environment and conducted a series of experiments on benchmark JSSP instances ~\cite{FT, LA, TA}, comparing its scheduling performance against traditional PDRs.

In the first experiment, we evaluate the framework on the classical JSSP to verify its implementation and validate its capability for training RL agents in job shop scheduling. An RL-based scheduling approach is implemented using Stable Baselines3~\cite{stbaselines} implementation of the PPO algorithm. The agent's action space consists of a binary decision: whether to execute the next scheduling action or defer it. The observation space is defined by seven features representing the current system state and the next available scheduling action. To ensure the agent learns scheduling strategies purely based on fundamental system dynamics, it is trained with a minimal reward function, a simplified action space, and a compact observation space, avoiding excessive domain-specific tuning.

For validation, we focus on comparing our RL agent with traditional PDRs, as these methods offer scalability and serve as reliable baselines, especially when real-world constraints are considered. Exact methods such as Constraint Programming or Mixed-Integer Linear Programming struggle to efficiently solve large, complex instances within reasonable time frames, making them impractical for comparison against an agent trained in a real-world framework. By prioritizing makespan minimization in our experiments, we provide a clear and direct benchmark between the RL-based approach and established heuristics.

Although our framework natively supports multi-objective optimization, this study primarily benchmarks performance on the makespan objective. This focus is motivated by the lack of widely accepted multi-objective benchmarks and standardized evaluation metrics for real-world JSSP. While classical single-objective benchmarks are well established, resources for multi-objective and real-world scenarios remain scarce. Developing such benchmarks and metrics is an important direction for future work, and our framework is designed to accommodate these evaluations as the field progresses.

\begin{figure}
    \includegraphics[width=1\linewidth]{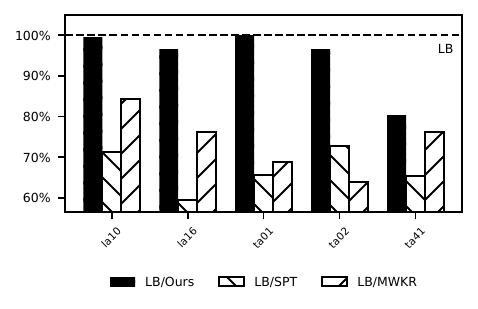}
    \vspace{-8mm}  
    \caption{Performance comparison of PDRs and RL agent in classical JSSP using the proposed framework}    
    \vspace{-5mm}  
    \label{Fig:Res1}
\end{figure}

Figure~\ref{Fig:Res1} illustrates the performance of our RL-based approach and of two traditional PDRs\textemdash{}Shortest Processing Time (SPT) and Most Work Remaining (MWKR)\textemdash{}across multiple benchmark JSSP instances. The figure presents the ratio of the known lower bound (LB) to the achieved makespan for each problem instance. Across all tested problems, the RL agent consistently achieves a makespan closer to the lower bound than the heuristic-based methods. This demonstrates that our framework effectively enables RL agents to learn optimized scheduling policies that outperform conventional rule-based approaches. These findings confirm the framework’s suitability for training RL agents on JSSPs and establish a solid foundation for extending the approach to more complex scheduling environments.

In the second experiment, we extended the classical JSSP by incorporating buffer and transport constraints to verify that our framework can represent real-world complexities and effectively train an RL agent under these conditions. To evaluate the RL-based scheduling approach in this setting, we compared its performance against the heuristic-based methods SPT and MWKR. Unlike the classical JSSP, where a known lower bound serves as a reference for optimality, the introduction of real-world constraints makes direct optimality comparisons impractical. Instead, the evaluation focuses on the relative performance improvements achieved by the RL agent over traditional heuristics.

\begin{figure}
    \includegraphics[width=1\linewidth]{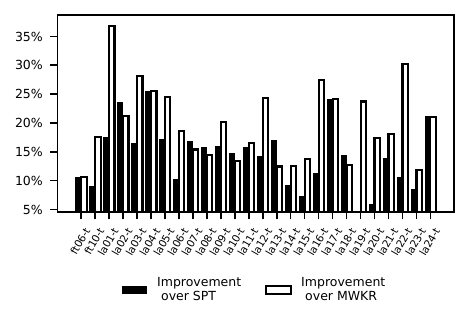}
    \caption{Performance improvements of RL agents over selected PDRs in our framework with real-world constraints}    
    \vspace{-5mm}  
    \label{Fig:Res2}

\end{figure}

Figure~\ref{Fig:Res2} presents the relative improvement in makespan achieved by our RL approach compared to SPT and MWKR across multiple extended JSSP instances. The results demonstrate that, despite the added complexity, the RL agent consistently outperforms both PDRs, learning to adapt its scheduling decisions to account for transport and buffer constraints more effectively than rule-based methods. These findings confirm that our framework successfully models real-world constraints and enables the training of RL agents capable of optimizing scheduling decisions in dynamic production environments.
This directly aligns with the core objective of our work: developing an RL-based scheduling framework capable of handling complex real-world constraints that traditional methods struggle to address. 

\section{CONCLUSIONS}
By extending classical JSSP with real-world constraints such as transport logistics, buffer management, machine breakdowns, and sequence-dependent setup times, we developed a framework that mirrors the complexities of modern manufacturing systems. Its modular design allows for tailored simulations of various production environments, while the unified interface enables the configuration of custom action spaces, observation structures, and reward functions for RL agents. This flexibility supports the development and evaluation of novel RL strategies for diverse scheduling challenges.
The framework enables an adaptive, data-driven approach for optimizing scheduling decisions in environments too complex for exact solvers, thereby making RL a viable alternative for industrial scheduling applications. It is particularly relevant because it addresses the lack of a comprehensive framework for standardized comparison and evaluation of scheduling approaches in realistic production environments.

Future work will refine the framework by investigating diverse action spaces, observation structures, and reward functions to optimize RL-based scheduling. By adjusting environmental complexity, we aim to achieve the optimal balance for integrating RL agents into real-world production. Another direction involves integrating the framework with real-time production systems, enabling adaptive online scheduling under dynamic constraints. In addition, establishing standardized benchmarks and evaluation metrics for multi-objective scheduling remains a central goal. Leveraging the framework’s extensibility, we seek to close the gap between academic research and industrial requirements. These developments will enhance robustness, scalability, and increase the framework’s technology readiness level, supporting real-world deployment. 

\addtolength{\textheight}{-10cm}   








\end{document}